\pdfoutput=1
\documentclass[11pt]{article}

\usepackage[final]{acl}

\usepackage{times}
\usepackage{latexsym}

\usepackage{multirow}
\usepackage{adjustbox}
\usepackage{wrapfig}
\usepackage{kotex}
\usepackage{enumitem}
\usepackage{amsmath}
\usepackage{lipsum}
\usepackage{blindtext}
\usepackage{multirow}
\usepackage{bbding}

\usepackage[T1]{fontenc}
\usepackage[utf8]{inputenc}

\usepackage{microtype}

\usepackage{inconsolata}

\usepackage{graphicx}
\usepackage{bbm}

\title{Dynamic Uncertainty Ranking: Enhancing Retrieval-Augmented In-Context Learning for Long-Tail Knowledge in LLMs}

\author{Shuyang Yu\textsuperscript{1}\thanks{Work was done during the internship at GE Healthcare.}, 
        Runxue Bao\textsuperscript{2}\thanks{Correspondence to: Runxue Bao, Cao Xiao <\{runxue.bao, cao.xiao\}@gehealthcare.com>}, 
        Parminder Bhatia\textsuperscript{2},\\ 
        \bf Taha Kass-Hout\textsuperscript{2}, 
        Jiayu Zhou\textsuperscript{3}, 
        Cao Xiao\textsuperscript{2}\footnotemark[2]\\
        \textsuperscript{1}Department of Computer Science and Engineering, Michigan State University\\
        \textsuperscript{2}GE Healthcare \\
        \textsuperscript{3}School of Information, University of Michigan \\
       }

\usepackage{tablefootnote}
\usepackage{amsthm}
\usepackage{mathtools}
\usepackage{hyperref}
\usepackage[nameinlink,capitalize]{cleveref}
\hypersetup{colorlinks=true,linkcolor=blue,citecolor=blue,urlcolor=blue,pdfborder={0 0 0}}
\usepackage[normalem]{ulem} % for command \sout "strike out"
\usepackage{mathtools}
\usepackage{algorithm,algorithmic}

\usepackage[utf8]{inputenc} % allow utf-8 input
\usepackage[T1]{fontenc}    % use 8-bit T1 fonts
\usepackage{hyperref}       % hyperlinks
\usepackage{url}            % simple URL typesetting
\usepackage{booktabs}       % professional-quality tables
\usepackage{amsfonts}       % blackboard math symbols
\usepackage{nicefrac}       % compact symbols for 1/2, etc.
\usepackage{microtype}      % microtypography
\usepackage{color, colortbl}
\definecolor{Gray}{gray}{0.9}
\newcolumntype{g}{>{\columncolor{Gray}}c}

\usepackage{graphicx,wrapfig}
\usepackage{caption}
\usepackage{subcaption}
\usepackage{amsfonts}
\usepackage{url}
\usepackage{enumitem}
\usepackage{amsthm}
\usepackage{thmtools}
\usepackage{thm-restate}

\theoremstyle{definition}

\theoremstyle{remark}

\newcommand{\vb}{{\mathbf{b}}}

\newcommand{\vh}{{\mathbf{h}}}

\newcommand{\vW}{{\mathbf{W}}}

\newcommand{\cC}{{\mathcal{C}}}

\newcommand{\cE}{{\mathcal{E}}}

\newcommand{\cP}{{\mathcal{P}}}

\newcommand{\cT}{{\mathcal{T}}}

\newcommand{\cV}{{\mathcal{V}}}

\newcommand{\bc}{\begin{center}}
\newcommand{\ec}{\end{center}}

\newcommand{\bdm}{\begin{displaymath}}
\newcommand{\edm}{\end{displaymath}}

\newcommand{\beq}{\begin{equation}}
\newcommand{\eeq}{\end{equation}}

\newcommand{\bfl}{\begin{flushleft}}
\newcommand{\efl}{\end{flushleft}}

\newcommand{\bt}{\begin{tabbing}}
\newcommand{\et}{\end{tabbing}}

\newcommand{\beqn}{\begin{align}}
\newcommand{\eeqn}{\end{align}}

\newcommand{\beqs}{\begin{align*}} % no equation numbers
\newcommand{\eeqs}{\end{align*}}  % no equation numbers

\begin{document}
\maketitle
\begin{abstract}
%memorization of LLM-> long-tail knowledge
Large language models (LLMs) can learn vast amounts of knowledge from diverse domains during pre-training. However, long-tail knowledge from specialized domains is often scarce and underrepresented, rarely appearing in the models' memorization. Prior work has shown that in-context learning (ICL) with retriever augmentation can help LLMs better capture long-tail knowledge, reducing their reliance on pre-trained data.
Despite these advances, we observe that LLM predictions for long-tail questions remain uncertain
%highly sensitive 
to variations in retrieved samples. 
%Even slight changes in the retrieval set can mislead the prediction. 
To take advantage of the uncertainty in ICL for guiding LLM predictions toward correct answers on long-tail samples,
we propose a reinforcement learning-based dynamic uncertainty ranking method for retrieval-augmented ICL that accounts for the varying impact of each retrieved sample on LLM predictions. Our approach prioritizes more informative and stable samples while demoting misleading ones, updating rankings based on the feedback from the LLM w.r.t. each retrieved sample. To enhance training efficiency and reduce query costs, we introduce a learnable dynamic ranking threshold, adjusted when the model encounters negative prediction shifts.
Experimental results on various question-answering datasets from different domains show that our method outperforms the best baseline by $2.76\%$, with a notable $5.96\%$ boost in accuracy on long-tail questions that elude zero-shot inference. Our code is available at \url{https://github.com/Yu-shuyan/uncertian_ranker}.
\end{abstract}

\section{Introduction}
Pretrained large language models~\cite{brown2020language, touvron2023llama,almazrouei2023falcon} have achieved remarkable success across various natural language processing (NLP) tasks, such as summarization~\cite{zhang2019pretraining, van2024adapted}, question answering~\cite{jiang2021can,wang2024infuserki}, and code generation~\cite{li2023starcoder,wang2024unlocking}. These impressive results are largely due to their pre-training on vast, web-sourced datasets spanning multiple domains. However, these real-world datasets often follow a long-tail distribution~\cite{liu2019large,mallen2022not,dai2023long,sun2023head}, where knowledge from less frequent domains is underrepresented. Consequently, certain domain-specific information may be rarely or even never included in the LLMs' memorization~\cite{kandpal2023large}. As a result, LLMs struggle to provide accurate responses to queries drawn from these long-tail distributions, since the pre-training process fails to capture this sparse information.

\begin{figure*}
    \centering
\includegraphics[width=0.85\linewidth]{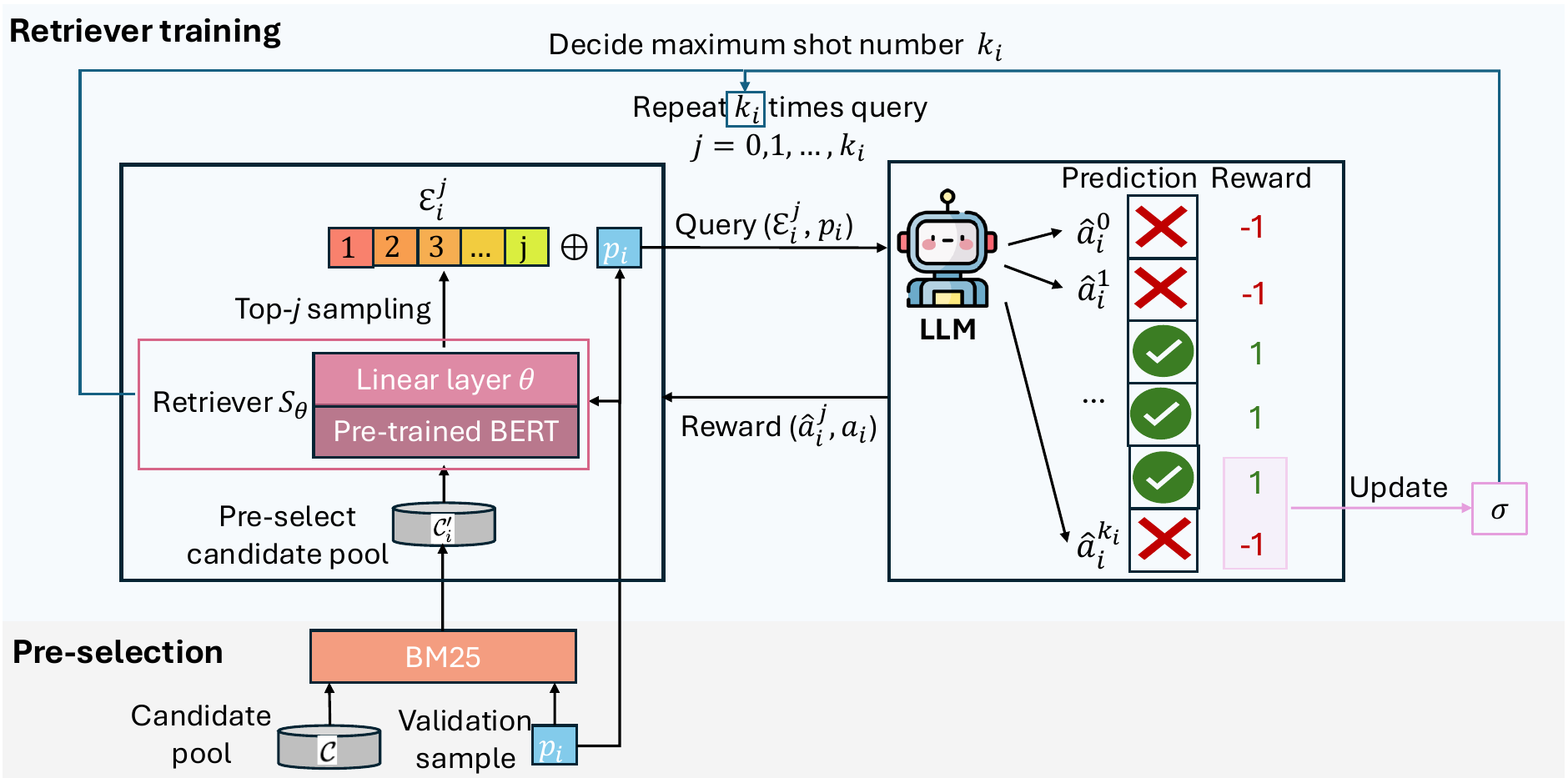}
    \caption{Training framework of the proposed method. After pre-selection using BM25 for each validation sample $p_i$, we conduct from $0$-shot to $k_i$-shot inference and update retriever $S_\theta$ according to the dynamic impacts of each sample on LLMs based on the reward from LLM. To reduce the query cost, we update the threshold $\sigma$ when the LLM experiences a negative prediction change. The query time $k_i$ is decided by retriever score $S_\theta$ and threshold $\sigma$. }
    \label{fig:framework}
 %        \vspace{-0.2in}
\end{figure*}

In-context learning (ICL)~\cite{brown2020language}  is a few-shot learning method that queries LLMs by concatenating relevant samples with the test query, without updating the model’s parameters.
\citet{kandpal2023large} found that ICL, when combined with retriever augmentation, can reduce LLMs' reliance on pre-training knowledge by retrieving relevant examples related to long-tail queries during inference. Common retrieval methods used to select augmentation examples for ICL include random selection~\cite{wei2022chain,wang2022self}, off-the-shelf retrievers (e.g., BM25~\cite{robertson2009probabilistic}), and fine-tuned retrievers (e.g., PromptPG~\cite{lu2022dynamic}).  However, prior works ~\cite{zhao2021calibrate,liu2021makes,lu2021fantastically,chen2023many} have shown that ICL with different selection and ordering of the retrieved samples could lead to unstable predictions of LLMs. 
%\cx{what do you mean "with different combinations of retrieved samples". This sentence needs more clarification}.
In our experiments, we observed a similar pattern: when utilizing existing methods to retrieve relevant samples for ICL, the model's predictions for long-tail questions—those not captured by zero-shot inference—exhibited particularly high uncertainty. In some cases, a subset of the retrieved samples led to correct predictions, while the full set misled the model, even with the same retrieval method.

In this paper, to enhance the retrieval augmentation for long-tail samples regarding LLM's uncertainty, 
%To address the challenge of LLMs' uncertain predictions, 
we propose a reinforcement learning-based dynamic uncertainty ranking method motivated by reinforcement learning's capacity to search for optimal retrieved samples based on the LLM's feedback~\cite{lu2022dynamic}. Specifically, our approach trains a retriever to prioritize informative and stable samples while down-ranking misleading ones, enhancing performance on both head and tail distributions. We build on the BERT-based retriever architecture~\cite{devlin2018bert} with an appended linear layer. During the training of the retriever, only the linear layer is fine-tuned. Initially, BM25~\cite{robertson2009probabilistic} is used for pre-selection, and the retriever is trained using policy gradients~\cite{sutton1999reinforcement}, guided by feedback from the LLM for each retrieved sample. To improve efficiency, we introduce a learnable dynamic threshold as a budget controller for retrieval, selecting only samples with high-ranking scores above this threshold, which adjusts whenever the LLM experiences a negative prediction change, i.e., the prediction changes from true to false. To evaluate the proposed approach, we compared our method with the state-of-the-art methods across both multi-choice and open-ended question-answering (QA) datasets from different domains. 
%including healthcare~\cite{jin2019pubmedqa}, speech detection~\cite{mollas2022ethos}, climate change~\cite{barbieri2020tweeteval}, and QA collected from Wikipedia~\cite{elsahar2018t,kwiatkowski2019natural} 
The experimental results show that our method outperforms the best baseline by $2.76\%$. Long-tail questions failed to be captured by a zero-shot inference benefit particularly from our proposed method. The accuracy of long-tail questions of our method
surpasses previous methods with a large margin of up to $5.96\%$.

%contributions
We summarize our key contributions as follows:
\begin{itemize}
\item We investigate the limitations of existing retrieval-augmented ICL approaches for handling long-tail questions, highlighting how variations in retrieved samples contribute to prediction uncertainty.
    \item We propose a reinforcement learning-based dynamic uncertainty ranking method with a budget controller that considers the dynamic impact of each retrieved sample on the LLM's prediction, which selectively elevates informative retrieved samples and suppresses misleading ones with minimal query costs.
    \item Extensive experiments  demonstrate that our method consistently outperforms the state-of-art method on multiple QA datasets from different domains, achieving nearly a $6\%$ improvement in accuracy for long-tail questions.
    %particularly improves LLM's prediction on long-tail questions that failed to be captured by a $0$-shot inference.
\end{itemize}

\section{Related Work}
\textbf{In-context learning (ICL).} ICL
\cite{brown2020language} queries the LLMs with a concatenation of related samples and the test query without parameter updating. To improve the quality of ICL, retrievers have been proposed to select related samples, which can be categorized into sparse retrievers (e.g. \cite{robertson2009probabilistic}) and dense retrievers (e.g. \cite{liu2021makes}). To further improve the effectiveness of the off-the-shelf retrievers, strategies for fine-tuning retrievers on specific target domains have been proposed such as PromptPG~\cite{lu2022dynamic}, UDR~\cite{li2023unified}, and LLM-R~\cite{wang2023learning}, etc. Some works also adopt GPT to help retrieve and rerank samples by providing special prompts and related samples, such as Rerank~\cite{sun2023chatgpt}, SuRe~\cite{kim2024sure}, etc.

\textbf{Long-tail knowledge learning for ICL.}
\citet{kandpal2023large} is the first to explore the influence of the long-tail distribution in pre-training data on LLM memorization.
%To better capture the long-tail knowledge, the first solution is model scaling. However, according to their experiments, models would need to be scaled dramatically to improve QA accuracy. Another more 
% promising approach is retrieval augmentation. 
 They find retrieval augmentation as a promising approach to significantly reduce the LLM's dependence on pre-training knowledge. Several subsequent works have built on this retrieval augmentation approach to address the long-tail problem in LLMs.
For example, \citet{dai2023long} propose a retrieve-then-rerank framework leveraging knowledge distillation (KD) from the LLM to tackle long-tail QA. However, their method involves tuning the language model, which is computationally expensive and impractical for black-box LLMs such as GPT-4~\cite{achiam2023gpt}. Another line of research focuses on augmenting the training set using GPT~\cite{saad2023udapdr,cloutier2023fine,li2023search}, followed by fine-tuning the retriever to enhance its performance.
%\cite{saad2023udapdr} used GPT and a less expensive LLM to generate large numbers of synthetic queries based on randomly sampled seed passages from the target domain, and then used the augmented training set to fine-tune the retriever. However, the random augmentation strategy is more likely to boost the number of head distribution samples, while the tail distribution samples are rarely augmented. 
%\cite{cloutier2023fine} proposed to augment human-like text using GPT according to the class to balance the dataset, but when encountering open-domain QA dataset, 
% Random augmentation strategy adopted by \cite{saad2023udapdr} is more likely to boost the number of head distribution samples, while the tail distribution samples are rarely augmented. 
% \cite{li2023search} adopted llama-7B~\cite{touvron2023llama} to approximate the tail distribution during augmentation, but the tail distribution of llama does not necessarily lie in the tail distribution of the inference model GPT due to their different memorization~\cite{kandpal2023large}. 
Nonetheless, determining which samples should be augmented remains challenging. Augmenting the training set based on seed sentences often introduces repetitive rather than diverse information, and incurs significant costs due to GPT queries. Therefore, in this paper, rather than augmenting the training set for fine-tuning the retriever, we aim to train an effective retriever capable of selecting the most informative samples to augment the test query during inference.
%solution2: augment the original candidate pool
%include augmentation for long-tail
%challenge1: do not know what kind of samples should be augmented. existing methods augment according to class, open questions do not have classes, we try clustering then augment, fail
%challenge2: existing methods augment according to the output logits of LLM, gpt4 is a black-box model, does not output logits
%challenge3: augment by paraphrasing based on several seed sentences, more repetitive information does not provide more information for inference prompt.
%challenge4: too much cost.
\section{Problem Formulation}
%in-context learning for question answering including multi-choices and open question answering
In this paper, we target in-context learning (ICL) for QA tasks including multiple-choice QA and open-ended QA from different domains. Suppose we have a training set $\cT = \{(x_i, y_i)\}_{i=1}^N$ related to the query domain, where $x$ is the question and $y$ is the answer. Given a query problem $p_i$ from a test set $\cP$ and a $K$-shot inference budget, we will retrieve $K$ related samples $\cE_i = \{e_i^k=(x_i,y_i)| e_i^k\in \cT\}_{k=1}^K$ and construct a prompt $P(\cE_i,p_i)$ as input to feed into the LLM:
\begin{align}
    P(\cE_i,p_i)= \pi(e_i^1)\oplus  \cdots \oplus \pi(e_i^K)\oplus\pi(p_i,\cdot),
\end{align}
where $\pi$ is the template for each sample. The predicted answer from the LLM for question $p_i$ is given by:
\begin{align}\label{eq:prediction}
    \hat{a}_i=\text{LLM}(P(\cE_i,p_i)).
\end{align}
%\rub{$p_i$ and $q_i$ is not consistent here. $q_i$ would be better if possible}

\begin{figure}
    \centering
    \includegraphics[width=1\linewidth]{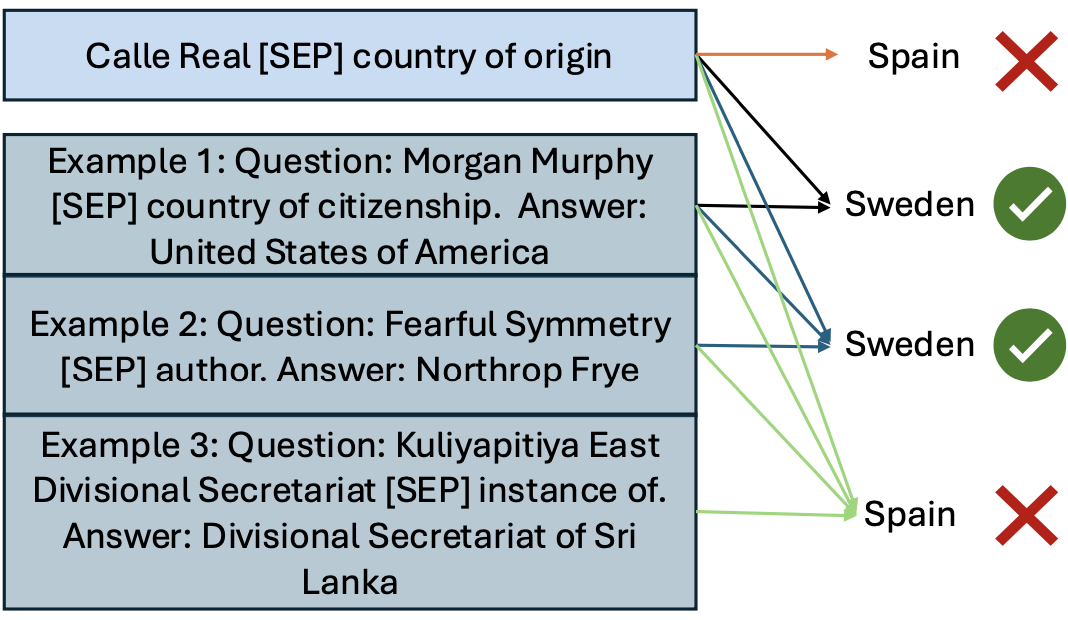}
         \vspace{-0.2in}
    \caption{Case study for uncertainty of ICL.}
    \label{fig:uncertain_case}
    \vspace{-0.2in}
\end{figure}
\begin{figure}
    \centering
    \includegraphics[width=1\linewidth]{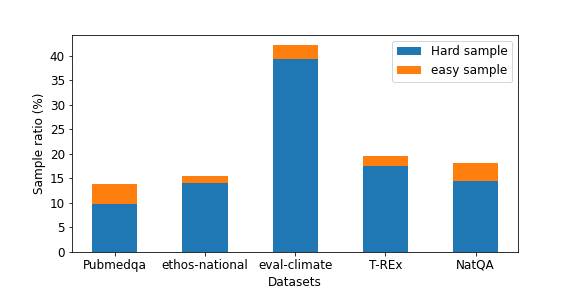}
     \vspace{-0.2in}
    \caption{Uncertain sample ratios.}
    \label{fig:uncertain_ratio}
    \vspace{-0.2in}
\end{figure}
\section{Motivation: Uncertainty of In-context Learning}\label{sec:motivation}
%easy hard samples based on the memorization of LLMs
%long-tail distribution knowledge the 0-shot LLM is unable to capture
%During the pre-training stage of LLMs, although the LLMs can learn a huge amount of information from web resources, 
Due to the lack of knowledge of some specific domains during the pre-training stage, there exists long-tail knowledge that failed to be captured by the 
LLMs~\cite{kandpal2023large}. 
%\rub{summarize this paragraph}
%uncertianty for easy/hard samples
%The LLMs are most uncertain about hard samples.
%\rub{Formally define with symbol} 
We define easy samples as queries that have been captured during the LLM's pre-training stage and are stored in its memorization. In contrast, hard samples refer to queries that the LLM failed to capture, which are more likely to represent long-tail data. We classify easy and hard samples using the zero-shot testing results $\hat{a}_i=\text{LLM}_{\text{0-shot}}(p_i)$:
\begin{align}\label{eq:easy_hard}
    &\cP_{\text{easy}}=\{(p_i,a_i)\in\cP|\mathbbm{1}(\hat{a}_i,a_i)=1\},\nonumber\\
    &\cP_{\text{hard}}=\{(p_i,a_i)\in\cP|\mathbbm{1}(\hat{a}_i,a_i)=-1\}，
\end{align} 
where the indicator function $\mathbbm{1}(\cdot)$ returns $1$ if the predicted answer $\hat{a}_i$ aligns with the ground truth answer $a_i$, otherwise it returns $-1$.
%Our goal in this paper is to improve the accuracy of these long-tail samples.
According to \citet{kandpal2023large}, retrieval augmentation methods help alleviate the long-tail problem, as when a retriever succeeds in finding the most relevant samples from the training set $\cT$, it reduces the LLM’s needs to have a large amount of related knowledge in its memorization. 
%uncertian for each retrieved sample given a certain budget, show answers can change from true to wrong
However, our experiments revealed that the LLMs exhibit higher uncertainty when presented with hard samples, regardless of the retrieval augmentation applied.
\cref{fig:uncertain_ratio} shows the uncertain sample ratios that experienced a prediction change on five datasets.
%As shown in \cref{fig:uncertain_ratio}, 
%During testing of questions from dataset T-REx~\cite{elsahar2018t},
Given a certain inference budget $K=5$, $21.84
\%$ of queries experience a prediction change when we increase from $0$-shot to $5$-shot. Among these uncertain queries, $87.18\%$ are hard samples and $12.82\%$ samples are easy samples using BM25 retrieval~\cite{
robertson2009probabilistic}. 
%Similar uncertain patterns of queries are also observed for other datasets and retriever methods as shown in our experiments (\textcolor{red}{add experiment section later}). 
For hard samples,
even a tiny variation in retrieved set $\cE$ can mislead the LLM's prediction. One case study for hard sample queries from T-REx~\cite{elsahar2018t} is shown in \cref{fig:uncertain_case}. In this case, LLM gives a correct answer with the first two informative samples in $\cE$, effectively compensating for the LLM's long-tail knowledge.
However, the answer gets wrong when a third sample is added to the prompt, which indicates the newly added knowledge is misleading. Other cases to show the uncertain prediction of LLM can be found in \cref{fig:hard_case} in \cref{sec:case} and \cref{tab:extended_uncertian_case} in Appendix.

% Our goal is to maintain the accuracy for 
%easy samples while improving the accuracy for hard samples, increase their stability
%for easy samples: martian performance, do not include misleading knowledge
%for hard samples:need knowledge compensation->in-context samples
Given the uncertainty of in-context learning, our goal is to improve the prediction accuracy of hard samples while maintaining the prediction stability on easy samples. 
%challenge: do not have prior knowledge of which sample is easy/hard 
During testing, we lack prior knowledge to determine whether a query falls into the easy or hard category. The primary challenge, therefore, is to prevent the inclusion of misleading information in the retrieved set $\cE$, which could lead to incorrect predictions. Simultaneously, we must ensure that the retrieved samples are sufficiently informative to address long-tail knowledge gaps and guide the LLM toward the correct answer.

\section{In-context Learning with Dynamic Uncertainty Ranking}
%\rub{include long-tail in this section. eg, when $j=0$, we conduct a long-tail detection}
In this section, we introduce a dynamic uncertainty ranking method built on a reinforcement learning-based retriever. This method adjusts the retriever by applying a dynamic threshold, lowering the rankings of misleading samples while elevating the rankings of informative and stable ones.

\subsection{Retrieved Sample Selection}
The original training set $\cT$ is randomly divided into a validation set $\cV$, and
a candidate pool $\cC$, from which the retrieved sample set $\cE$ is selected. Following \citet{lu2022dynamic}, the retriever structure is built upon BERT~\cite{devlin2018bert} with a linear layer appended to the final pooling layer of the BERT model. During training, the BERT is frozen, and only the parameter $\theta=(\vW,\vb)$ of the linear layer is fine-tuned. Given a query $p_i$ from the validation set $\cV$ and a retrieved sample $e_i$ from $\cC$, the ranking score of the retriever is achieved by the hidden logical similarity shared among samples:
\begin{align}\label{eq:score}
    S_{\theta}(e_i|p_i)=\frac{\exp[\vh (e_i)\cdot \vh(p_i)]}{\sum _{e_i'\in \cE}\exp [\vh (e_i')\cdot \vh(p_i)] },
\end{align}
where $\vh(\cdot)=\vW(\text{BERT}(\cdot))+\vb$ is the output of the linear layer. 

To ensure the diversity and similarity of retrieved samples, and reduce the computational cost,
we first adopt an off-the-shelf retriever BM25~\cite{robertson2009probabilistic} to pre-select a small candidate set $\cC_i'$ from the large candidate pool $\cC$ following~\citet{rubin2021learning,sun2023chatgpt,kim2024sure}. 

Suppose the shot number is $k$, 
by selecting samples with the Top-$k$ highest ranking score using our retriever $S_\theta$, we can achieve the retrieved sample set $\cE_i$ for $p_i$ from candidate pool $\cC_i'$ as follows:
\begin{align}
    \cE_i = \{e_i^k \sim \text{Top-}k(S_{\theta}(e_i^k|p_i))|e_i^k\in \cC_i'\}. 
\end{align}

The retriever selection process for testing is the same as the training, the only difference is the validation set $\cV$ will be replaced with the test set $\cP$.

\subsection{Retriever Training}
%To enhance the retrieval augmentation for long-tail samples,
%According to our findings in \cref{sec:motivation}, uncertain predictions are common for LLMs, and even a tiny fluctuation of the retrieved sample set can mislead the prediction. 
%To better handle the uncertainty of LLMs, 
%we propose a dynamic ranking method that updates the retriever based on the dynamic feedback from LLM 
%with different retrieved samples based on the investigation of LLM's uncertainty in \cref{sec:motivation}.
Motivated by the exploration in \cref{sec:motivation},
to improve retrieval augmentation for both hard and easy samples, we introduce a dynamic ranking method that updates the retriever using feedback from the LLM, driven by its varying responses to each retrieved sample.
%\rub{first time to mention head?}

\textbf{Decide maximum shot number.} Before training, we first decide the maximum shot number for each validation sample $p_i\in\cV$.
To achieve this, we define a maximum shot number budget $K$ and a dynamic budget controller $\sigma$ initialized as $0$ for ranking scores $S_\theta$. Only samples with ranking scores above the threshold $\sigma$ will be selected to update the retriever.
The maximum shot number $k_i$ for $p_i$ is: 
%decided by:
\begin{align}\label{eq:shot_number}
    k_i=\min(K, N_i^{\max}),
    %N_i^{\max} = | \{e_i^k \sim S_{\theta}(e_i^k|p_i)|e_i^k\in \cC_i', S_{\theta}(e_i^k|p_i)>\sigma\}|.
\end{align}
where $N_i^{\max} = | \{e_i^k \sim S_{\theta}(e_i^k|p_i)|e_i^k\in \cC_i', S_{\theta}(e_i^k|p_i)>\sigma\}|$.

\textbf{Training process.} Given the maximum shot number $k_i$, 
we then conduct inference for $p_i$ from $0$-shot to $k_i$-shot to capture the effect of each retrieved sample on the LLM. The $0$-shot inference on $p_i$ can be considered as a means of long-tail sample detection as defined in \cref{eq:easy_hard}. If the model's answer is incorrect, the sample is classified as a hard sample (i.e., long-tail sample), and the retrieved set should provide informative augmentation. Conversely, if the model produces the correct answer, the sample is classified as an easy sample, and the retrieved set should avoid introducing any misleading samples.
We define the retrieved sample set for the $j$-shot inference as the top-$j$ highest ranking score selected from candidate pool $\cC_i'$:
\begin{align}\label{eq:E_i}
     &\cE_i^j = \{e_i^k \sim \text{Top-}j(S_{\theta}(e_i^k|p_i))|e_i^k\in \cC_i'\},\\ &\text{where}\quad j=\{0,1,\cdots,k_i\}.\nonumber 
\end{align}
The prediction from LLM based on $\cE_i^j$ and $p_i$ is generated according to \cref{eq:prediction} as $\hat{a}_i^j=\text{LLM}(P(\cE_i^j,p_i))$. The retrieved sample's impact on the prediction is reflected by the reward function $ R(\hat{a}_i^j,a_i)= \mathbbm{1}(\hat{a}_i^j, a_i)$,
where $a_i$ is the ground truth answer for $p_i$, $\mathbbm{1}(\cdot)$ is the  indicator function.

 Our training goal is to maximize the expected reward w.r.t. the parameters of the retriever using the Policy Gradient method~\cite{sutton1999reinforcement}. Since the expected reward cannot be computed in closed form, following~\citet{lu2022dynamic}, we compute an unbiased estimation with Monte Carlo Sampling:
\begin{align}\label{eq:reward_obj}
    \!\!\!\mathbbm{E}_{e_i \sim S_\theta (e_i|p_i)}[(\hat{a}_i, a_i)]
    \approx \frac{1}{N}\sum_{i=1}^N\sum_{j=1}^{k_i}R(\hat{a}_i^j, a_i),
\end{align}
where $N$ is the batch number yielded from $\cV$. Following the REINFORCE policy gradient~\cite{williams1992simple}, we update the retriever using:
\begin{align}\label{eq:update_loss}
   \nabla  &\mathbbm{E}_{e_i \sim S_\theta (e_i|p_i)}[R(\hat{a}_i, a_i)] \nonumber\\
   =& \mathbbm{E}_{e_i \sim S_\theta (e_i|p_i)} \nabla_\theta \log (S_\theta (e_i|p_i))R(\hat{a}_i, a_i)\nonumber\\
   \approx&\frac{1}{N}\sum_{i=1}^N\sum_{j=1}^{k_i}\nabla_\theta \log(S_\theta (e_i^j|p_i))R(\hat{a}_i^j, a_i),
\end{align}
%\rub{change $R(\text{LLM}(P(e_i,p_i)), a_i)$ to $\hat{a_i}$}
where $e_i^j=\cE_i^j - \cE_i^{j-1}$ is the difference between the retrieved sets for $j$-shot and $(j-1)$-shot. This approach incorporates the dynamic influence of each retrieved sample on the LLM, providing a better handling of uncertainty in ICL. Specifically, retrieved samples that yield correct predictions ($R(\cdot)=1$) are treated as informative and contribute to augmenting long-tail knowledge, thus receiving a higher ranking. Conversely, retrieved samples that lead to incorrect predictions ($R(\cdot)=-1$) are considered misleading and are ranked lower.

\begin{algorithm}[thb]
   \caption{ICL with dynamic uncertainty ranking}
   \label{algorithm:dynamic_rank}
\begin{algorithmic}[1]
   \STATE \textbf{Input}: Retriever $S_\theta$, training set $\cT$, maximum shot number $K$.
   %\# of training epochs $E$, 
   %threshold $\sigma$ \\
   \STATE \textbf{Output}: Trained retriever $S_\theta$ .
   % \STATE Randomly split $\cT$ into a validation set $\cV$ and a candidate pool $\cC$.
   \STATE Randomly split $\cT$ into $\cV$ and  $\cC$.
   \STATE Initialize $\theta \leftarrow\theta_0$, threshold $\sigma \leftarrow 0$.
   %\FOR{epoch=$1,2,\cdots,E$}
   \FOR{$\cV_{\text{batch}} \in \cV$}
   \STATE Initialize batch loss $L\leftarrow 0$.
   \FOR{each validation sample $p_i \in \cV_{\text{batch}}$}
   \STATE Pre-select $\cC_i'$ from $\cC$ using BM25 for $p_i$.
   \STATE Calculate the maximum shot number $k_i$ based on $\sigma$ using \cref{eq:shot_number}.
   \FOR{$j=0, 1,\cdots,k_i$}
   \STATE Get the retrieved set $\cE_i^j$ using \cref{eq:E_i}.
   % \STATE Conduct $j$-shot inference for $p_i$ with the retrieved set $\cE_i^j$.
   \STATE Get prediction  $\hat{a}_i^j=\text{LLM}(P(\cE_i^j,p_i))$.
   \STATE Get reward $R(\hat{a}_i^j,a_i)= \mathbbm{1}(\hat{a}_i^j, a_i)$.
   \STATE $L\leftarrow L -  R(\hat{a}_i^j,a_i)\cdot \log (S_\theta (e_i^j|p_i))$.
   \IF{$R(\hat{a}_i^{j},a_i)=-1, R(\hat{a}_i^{j-1},a_i)=1$}
    %\STATE $\sigma \leftarrow \max(S_\theta(e_i^k|p_i))$, where $e_i^k \in \cE_i^{k_i}-\cE_i^{j-1}$.
     \STATE Update $\sigma$ using \cref{eq:update_sigma}.
   \ENDIF
   \ENDFOR

   \ENDFOR
   \STATE Optimize $L$ w.r.t. $\theta$ using \cref{eq:update_loss}.
   \ENDFOR
   
   %\ENDFOR

\end{algorithmic}
\end{algorithm}

\begin{table*}[htbp!]
    \centering
    \small
    \scalebox{0.95}{
    \begin{tabular}{lccccccc}
    \toprule
         \multirow{2}{*}{Retrieval Method}&\multicolumn{4}{c}{Dataset}&&\multirow{2}{*}{Avg}\\
         \cline{2-6}&Pubmedqa&ethos-national&eval-climate&T-REx&NatQA\\ \hline \hline
         0-shot&$72.87\pm0.31$&$75.61\pm0.51$&$46.30\pm 0.32$&$42.60\pm2.36$&$44.20\pm1.91$&$56.32\pm 1.08$\\
         Random sampling&$78.20\pm 0.53$&$75.17\pm 1.01$&$66.30\pm3.53$&$57.13\pm1.97$&$46.80\pm1.44$&$64.72\pm 1.70$ \\
         BM25&$78.93\pm0.31$&$87.47\pm 0.39$&$82.57\pm0.30$&$62.13\pm1.33$&$55.00\pm1.14$&$73.22\pm 0.69$ \\
         SuRe&$78.93\pm0.42$ &$85.23\pm 0.33$&$78.89\pm0.30$&$39.80\pm0.57$&$32.00\pm3.40$&$62.97\pm1.00$\\
         Rerank&$78.93\pm0.42$ &$89.15\pm 0.39$ & $83.22\pm0.32$&$62.07\pm2.01$ &$53.80\pm 1.91$&$73.43\pm1.01$ \\
         PromptPG&$78.47\pm0.90$&$77.74\pm 2.16$&$72.78\pm2.00$&$60.73\pm3.21$&$50.80\pm 2.00$&$68.10\pm 2.05$  \\
         \hline     Ours&$\boldsymbol{80.60\pm0.35}$&$\boldsymbol{92.40\pm0.20}$&$\boldsymbol{85.37\pm0.32}$&$\boldsymbol{65.00\pm2.69}$&$\boldsymbol{57.60\pm1.91}$&$\boldsymbol{76.19\pm 1.09}$ \\
         \bottomrule
   \end{tabular}
   }
    \caption{Comparison results between proposed methods and baselines on QA tasks from different domains.}
    \vspace{-15pt}
    \label{tab:main}
\end{table*}
\begin{figure*}
    \centering
    \begin{subfigure}[b]{0.49\textwidth}
        \centering
        \includegraphics[width=1\textwidth]{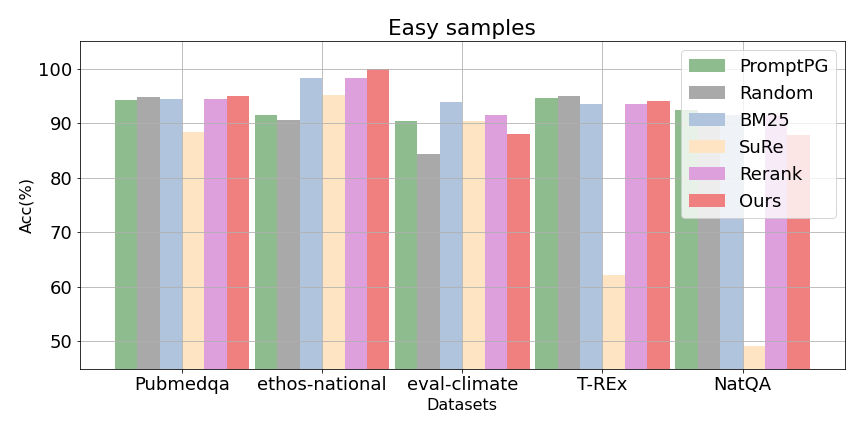}
        \vspace{-15pt}
        \subcaption{Easy sample accuracy.}
    \end{subfigure}
    \begin{subfigure}[b]{0.49\textwidth}
        \centering
        \includegraphics[width=1\textwidth]{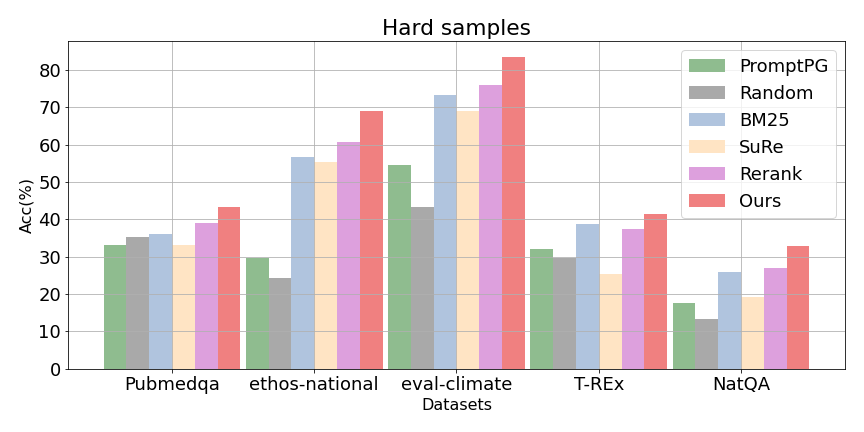}
        \vspace{-15pt}
        \subcaption{Hard sample accuracy.}
    \end{subfigure}
    \caption{Accuracy on easy and hard samples for proposed method and baselines.}
    \vspace{-15pt}
    \label{fig:easy_hard}
\end{figure*}

\textbf{Update budget controller $\sigma$.} 
In order to increase training efficiency and reduce the cost of querying the LLM, we also update the threshold $\sigma$ that served as a budget controller at the turning point for prediction change to decrease the inference times while maintaining the effect of our training strategy.
Specifically, we focus on a special case: when
%to pay more attention to the long-tail knowledge compensation during training.
the LLM experiences a prediction change from true to false， i.e., $R(\hat{a}_i^{j-1},a_i)=1$ and $R(\hat{a}_i^{j},a_i)=-1$. In this case, the first $(j-1)$-th samples have a positive impact on the inference of LLM, while the $j$-th sample has a negative impact.
%In this case, the first $(j-1)$-th samples can be considered as informative samples to augment long-tail knowledge, while the $j$-th sample can be considered as a misleading sample. Thus, the first $(j-1)$-th samples should be given higher rankings, and the $j$-th sample should be given a lower ranking instead.
Thus, 
 we update the threshold $\sigma$ as the maximum value of the ranking score for unselected samples in $\cE_i^{k_i}$ for the $(j-1)$-shot round as follows:
 \begin{align}\label{eq:update_sigma}
     \sigma = \max(S_\theta(e_i^k|p_i)),\quad
     e_i^k \in \cE_i^{k_i}-\cE_i^{j-1}.
 \end{align}
 Since we only select samples with ranking scores larger than $\sigma$ as shown in \cref{eq:shot_number}, the retrieved samples that serve as a good compensation for long-tail knowledge will be ranked higher, and be used for updating the retriever more frequently. Note that updating $\sigma$ will not wipe out the updating of misleading samples, as the turning point for prediction change is different for each validation sample. Without affecting our original training strategy, we improve the efficiency and deduct the querying cost. Our algorithm is summarized in \cref{algorithm:dynamic_rank}.

\section{Experiments}
In this section, we first introduce the experiment setup and then show the effectiveness of our method through various empirical results.
\subsection{Experimental Setup}
\textbf{Datasets:} We conduct the experiments on QA datasets from different domains, including
 three multi-choice datasets: biomedical dataset \emph{Pubmedqa}~\cite{jin2019pubmedqa}, speech detection dataset \emph{ethos-national}~\cite{mollas2022ethos}, climate change dataset \emph{eval-climate
}~\cite{barbieri2020tweeteval}, and two open-ended QA dataset: \emph{T-REx}~\cite{elsahar2018t} and  NaturalQuestions (\emph{NatQA})~\cite{kwiatkowski2019natural}. More datasets details can be found in \cref{sec:exp_detail}.

\textbf{Baselines:} We compare our method with six baselines, including \emph{0-shot} inference and five few-shot retrieval augmentation methods. The retrieval augmentation methods are as follows: 1) \emph{Random sampling}: selecting ICL samples from the candidate set, a widely adopted practice in many ICL studies~\cite{wei2022chain,wang2022self}; 2) \emph{BM25}~\cite{robertson2009probabilistic}: an off-the-shelf sparse retriever; 3）SuRe~\cite{kim2024sure}: first use GPT to summarize the retrieved passages from BM25 for  multiple answer candidates, then determines the most plausible answer by evaluating and ranking the generated summaries; 4) Rerank~\cite{sun2023chatgpt}: use GPT to rerank samples retrieved by BM25;
5) \emph{PromptPG}~\cite{lu2022dynamic}: a BERT-based dense retriever trained using reinforcement learning based on the feedback from GPT.

\textbf{Evaluation:} For multi-choice QA, we use accuracy for evaluation. For open-ended QA, we use Normalized Exact Match (NEM), which evaluates whether the normalized string output by the inference LLM is identical to the reference string.

\textbf{Implementation:} The LLM used in our experiment is GPT-4~\cite{achiam2023gpt}. Due to the limited data size in \emph{tweet\_eval-stance\_climate}, the training set is split into $50$ candidate samples and $150$ validation samples. For the other datasets, we use $1000$ samples in the candidate pool and $200$ samples in the validation set. All methods share the same train-test split. 
The number of pre-selected samples in $\cC'$ is set to $20$ by default for both the training and testing stages. For the few-shot case, the shot number is set to $5$, unless otherwise specified. During the training of our method, the maximum shot number budget $K$ is also set to $5$. The batch size is set to $20$. Experiments for all test datasets are repeated $3$ times with different seeds, and the average accuracy is reported in the results.
%The number of pre-selected samples in $\cC'$ is set to $20$ by default for both the training and testing stages. For the few-shot case, the shot number is set to $5$, unless otherwise specified. During the training of our method, the maximum shot number budget $K$ is also set to $5$. The batch size is set to $20$. Experiments for all test datasets are repeated $3$ times with different seeds, and the average accuracy is reported in the results.
\begin{figure*}[htbp!]
    \begin{minipage}{0.60\textwidth}
        \centering
    \begin{subfigure}[b]{0.48\textwidth}
        \centering
        \includegraphics[width=1\textwidth]{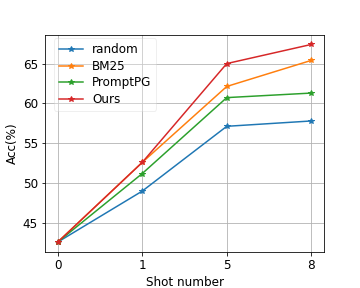}
        %\vspace{-10pt}
        \subcaption{T-REx.}
    \end{subfigure}
    \begin{subfigure}[b]{0.48\textwidth}
        \centering
        \includegraphics[width=1\textwidth]{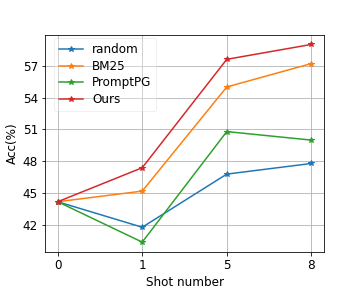}
        %\vspace{-10pt}
        \subcaption{NatQA.}
    \end{subfigure}
    \caption{Effects of different number of shots.}
    \label{fig:shots}
\end{minipage}
 \begin{minipage}{0.35\textwidth}
        \centering
        \hspace{6pt}
        \begin{subfigure}[b]{0.96\textwidth}
        \centering
        \includegraphics[width=1\textwidth]{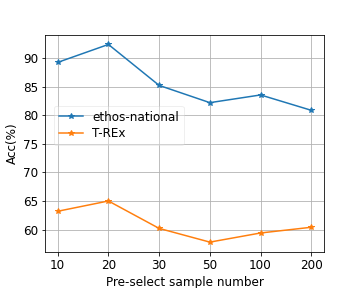}
        %\vspace{-10pt}
        %\subcaption{NatQA and ethos-national.}
    \end{subfigure}
    \caption{Effects of different pre-select numbers.
    }
    \label{fig:pre-select}
\end{minipage}
\end{figure*}
\subsection{Main Results}
%benchmark results on avg+std accuracy
 \cref{tab:main} presents the mean and standard deviation (std) of accuracy for our proposed method and the baselines across five QA datasets. Our approach outperformed all baselines across tasks, with an average improvement of $2.97\%$ 
ranging from $1.67\%$ to $3.25\%$ over the best baseline. The trained retriever PromptPG gives the most uncertain prediction with a std of $2.05\%$. Although our method is based on PromptPG, by 
%applying a sample-wise updating strategy and a dynamic threshold to 
giving informative and stable samples higher ranks, we not only improve the overall accuracy but also decrease std to $1.09\%$, comparable to 0-shot inference.
%Our improvement in certainty is especially significant for multi-choice QA %(PubmedQA, ethos-national, eval-climate) , we decrease the std to an average of $0.29\%$, which is even lower than the most stable baseline BM25.
%(1.69, 0.29, )
%\rub{can we place our method in figure 3 at the far right?}

We further investigate the accuracy of easy and hard samples in \cref{fig:easy_hard}. As illustrated in \cref{eq:easy_hard}, the easy/hard sample classification is decided by the 0-shot inference results, and the hard samples can be considered as long-tail questions of GPT-4.
% we define questions that failed to be captured by 0-shot inference for LLM as hard samples, and the successfully answered questions as easy samples. There is a high probability that hard samples are from the long-tail distribution of LLMs. 
First, we observe a similar pattern to ~\citet{kandpal2023large} that retrieval augmentation greatly improves the accuracy of long-tail samples. This could come from various aspects of augmented samples—such as label space, input text distribution, and sequence format—that collectively improve final predictions~\cite{min2022rethinking}. Compared with 0-shot inference, even random sampling improves accuracy on hard samples from $0\%$ to $29.17\%$. However, retrieval augmentation is highly dependent on the quality of the retrieval set. By retrieving the most similar samples, BM25 achieves an accuracy of $46.12\%$. Rerank further improves the accuracy to $48.03\%$.
%Due to the special selection strategy of our method, 
Our method includes the most informative samples based on the sample-wise feedback from LLM, and improves the accuracy on hard samples to $53.99\%$, which surpasses the best baseline with a large average
margin of $5.96\%$
ranging from $2.69\%$ to $8.11\%$, while maintaining the accuracy on easy samples. 
%as other baselines.

%accuracy on easy/hard samples

\subsection{Ablation Studies}
\begin{table}[]
    \centering
   \resizebox{0.48\textwidth}{!}{
    \begin{tabular}{ccccc}
    \toprule
         Dataset&PromptPG&UR&PromptPG+PS& UR+PS (Ours)\\
         \hline
         ethos-national&77.74&81.21&86.91&\textbf{92.40} \\
         Pubmedqa&78.47&80.10&79.10&\textbf{80.60} \\
         \bottomrule
    \end{tabular}
    }
    \caption{Effects of different components. PS denotes pre-selection. UR denotes uncertainty rank.}
    \label{tab:components}
    \vspace{-0.2in}
\end{table}
\textbf{Effects of different components.} We verify the effectiveness of two components of our proposed method: uncertainty rank  and pre-selection in \cref{tab:components}. We first compared the uncertainty rank (UR) strategy with another trained retriever PromptPG which shared the same retriever architecture as ours. We improve the accuracy by $3.47\%$ and $1.63\%$ for two different datasets. PromptPG adjusts the ranking of candidate samples based on the feedback on the entire retrieved set for the validation samples, while UR 
raises the ranks for informative and stable samples and lowers the ranks for misleading samples based on the sample-wise feedback from LLMs. 
%dynamically with a budget controller. 
UR avoids the condition when misleading samples are included 
%in the retrieved set 
and negatively changes the answer from true to false. In this way, UR greatly enhances the retrieved sample set for augmentation. 

The second component pre-selection (PS) improves the results of both PromptPG and UR by selecting more diverse and similar related samples in the candidate set $\cC'$. Then the second step retrieval can select samples from a smaller candidate pool of higher quality. By combining these two components together, we can achieve an overall improvement of $14.66\%$ and $2.13\%$ for two different datasets.
The improvement on ethos-national is more significant than Pubmedqa because the predicted answer on ethos-national is more uncertain given different combinations of retrieved samples. 

\begin{figure*}[]
    \centering
    \includegraphics[width=1\linewidth]{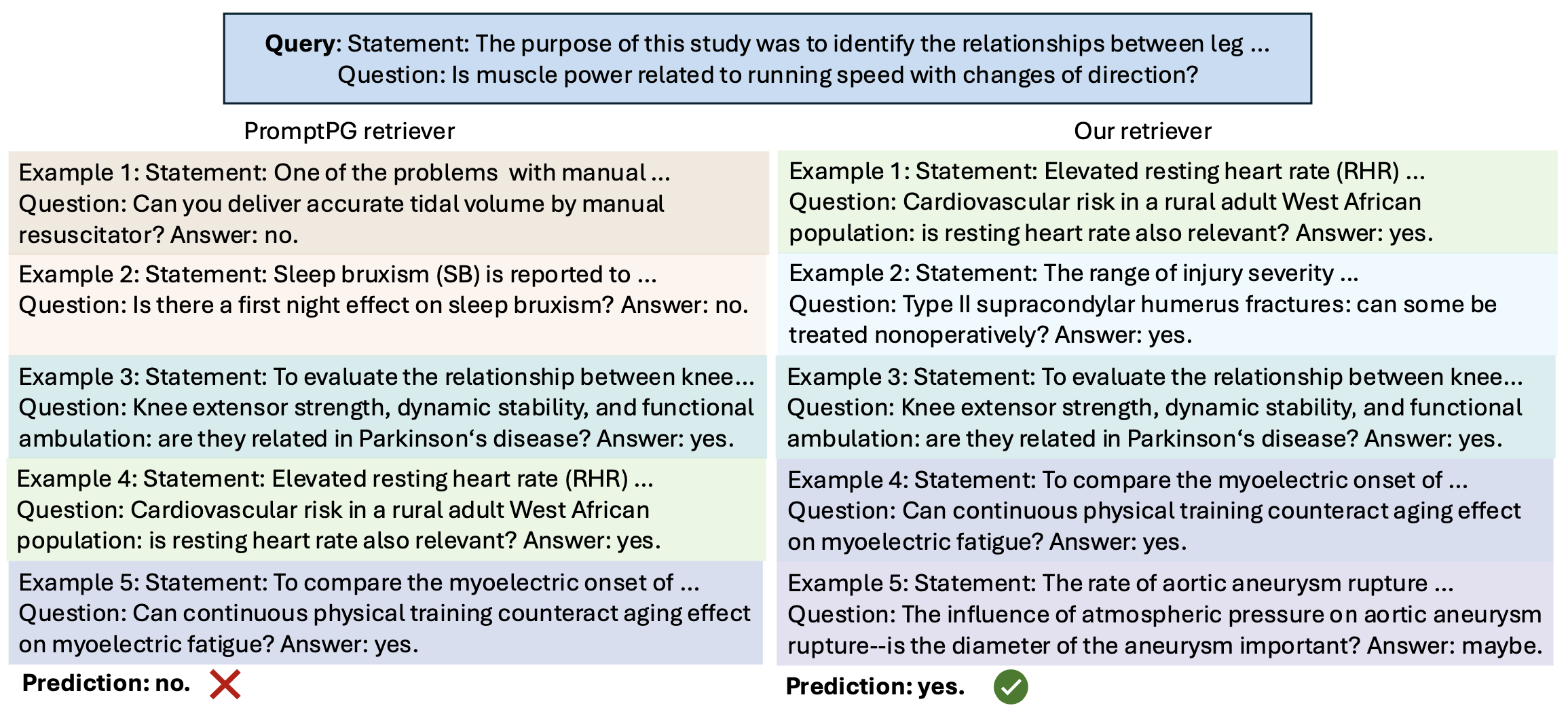}
    \caption{Case study for retrieved samples of hard samples.}
    \label{fig:hard_case}
    \vspace{-0.2in}
\end{figure*}

%\rub{can we use another dataset in Figure 5 to better showcase the argument}
\textbf{Effects of different number of shots.}
We show the effects of different shot numbers for two datasets in \cref{fig:shots} where our method consistently outperforms other baselines. 
%For T-REx, more shots in the retrieved set augment more information to the test query leading to better prediction results. 
For NatQA, the accuracy of random sampling and PromptPG retrieval does not monotonically increase with shot number due to low-quality, misleading samples, which can degrade performance. In contrast, our method prioritizes high-quality samples, and as the number of shots increases, the advantages of our algorithm become more pronounced, resulting in improved accuracy.

\textbf{Effects of different number of pre-selection samples.} In \cref{fig:pre-select}, we investigate how the number of pre-seletion samples impacts our algorithm. For both datasets, the accuracy first increases and then decreases. If too few samples are selected, the candidate pool $\cC'$ for our reinforcement learning
-based ranking stage lacks diversity, limiting the policy gradient strategy's action space. Consequently, the learned retriever struggles to find the most informative samples. If the number is too large, $\cC'$ includes many irrelevant samples, making it difficult for the policy gradient strategy to learn an optimal solution in the large search space~\cite{lu2022dynamic}. This can lead the retriever to capture irrelevant or misleading information.

\subsection{Case Study}\label{sec:case}
%\textbf{Case study for uncertain samples}
%\textbf{Case study for retrieved samples of hard samples}
To intuitively show the effectiveness of our proposed method on hard samples, we show one case on Pubmedqa by comparing the retrieved samples of PromptPG retriever and our retriever in \cref{fig:hard_case}. According to this case, the two retrieved sets even have three overlap samples (marked as the same color), but the prediction is completely different. PromptPG gives a wrong prediction answer, while our method delivers the right answer. This result verifies that GPT-4 gives uncertain predictions on long-tail samples. Since $0$-shot inference gives a wrong prediction answer on this query question, the informative augmented information can be contained in the retrieved set of our method (see right column),
while for PromptPG, misleading 
information can be contained in the two samples that do not intersect with our retriever set (see left column), which shifts the predicted answer from true to false. Compared with PromptPG, our retriever ranks the three overlapped samples higher and gives two more informative samples. With the combination effect of these two, our method gives the correct prediction. More cases on hard samples from other datasets can be found in \cref{tab:hard_case_trex} in the appendix.

\subsection{Efficiency Analysis}
\begin{figure}[htbp!]
    \centering
     \begin{subfigure}[b]{0.235\textwidth}
        \centering
        \includegraphics[width=1\textwidth]{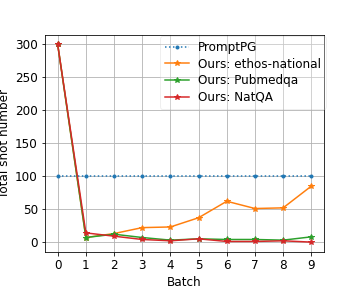}
        \vspace{-10pt}
        \subcaption{Shot \# w.r.t. the batch.}
        \label{fig:efficiency_line}
    \end{subfigure}
     \begin{subfigure}[b]{0.235\textwidth}
        \centering
        \includegraphics[width=1\textwidth]{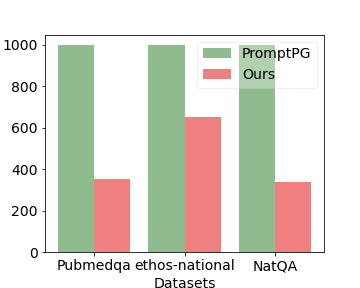}
        \vspace{-10pt}
        \subcaption{Shot \# of one epoch.}
        \label{fig:efficiency_bar}
    \end{subfigure}
    \caption{Efficiency analysis.}
    \label{fig:efficiency}
    \vspace{-0.2in}
\end{figure}

\textbf{Query cost.}
We set threshold $\sigma$ as the budget controller to reduce the cost of the querying GPT-4. Since the query cost depends on token length, we compare the query costs of our method and PromptPG (both trained based on GPT-4)
in \cref{fig:efficiency_line}. Specifically, we calculate the total number of shots included in each query during training for each batch within one epoch for both methods. The blue dash line shows the total shot number of PromptPG for all datasets, since the batch size is $20$, and the shot number is fixed at $5$, the total shot number is fixed at $100$ for each batch. 
According to the results, only batch $0$ of our method surpasses PromptPG with a total shot count of $300$. For subsequent batches, as the threshold $\sigma$ is adjusted based on changes in the LLM's predictions, the query shot count drops significantly, resulting in the total shot count consistently being lower than that of PromptPG.
Aggregating the shot numbers across $10$ batches,
our method achieves only $33.8\%$, $65.2\%$, and $35.3\%$ of the shot count of PromptPG on Pubmedqa, ethos-national, and NatQA, respectively as shown in \cref{fig:efficiency_bar}. Thus, in conjunction with the accuracy comparison presented in \cref{tab:main}, our approach not only enhances query accuracy but also reduces the overall query cost.
%\rub{can we add a bar figure or a small table to show the average shot number reduction?}

\begin{figure}[htbp!]
    \centering
    \includegraphics[width=0.8\linewidth]{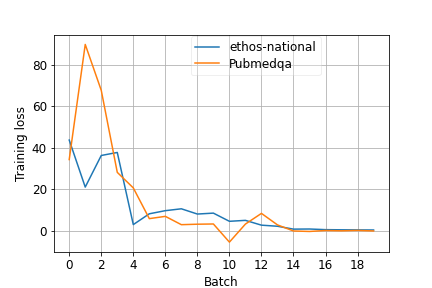}
    \caption{Training loss w.r.t. batch.}
    \label{fig:convergence}
\end{figure}

\textbf{Convergence speed.}  We empirically demonstrate the convergence speed by showing training loss curves in \cref{fig:convergence}. According to the results, the training loss quickly converges to a small value close to $0$ within $15$ batchs, which verify the high computational efficiency of our method. 

\subsection{Transferability Analysis} 
We investigate the transferability of our retriever in \cref{tab:transfer}. We use our retriever trained on dataset \emph{ethos-national}, and evaluate its cross-domain effectiveness across the rest of the four datasets.
Although the cross-domain results are still slightly inferior to the in-domain results, the performance gap is minimal, averaging only $0.98\%$. Furthermore, the cross-domain results outperform the best baseline. These findings indicate that our trained ranking strategy is transferable to other datasets, providing a cost-effective alternative to retraining.

 %\rub{can we include the the best baseline as shown in \cref{tab:main} in \cref{tab:transfer}}
\begin{table}[htbp!]
    \centering
    \resizebox{0.5\textwidth}{!}{
    \begin{tabular}{cccccc}
    \toprule
         & Pubmedqa&eval-climate&NatQA&T-REx&Avg \\
         \hline
          Best baseline&78.93&83.22&55.00&62.13&69.82\\
         Ours: cross-domain&79.60 &83.33&57.20&64.50&71.16 \\
         Ours: in-domain&80.60&85.37&57.60&65.00&72.14 \\
         \bottomrule
    \end{tabular}
    }
    \caption{Transferability of our method.}
    \label{tab:transfer}
\end{table}
\section{Conclusion}
 In this paper, to improve the uncertain prediction of LLMs on long-tail knowledge, we propose a reinforcement learning-based dynamic uncertainty ranking method for retrieval-augmented ICL with a budget controller. Specifically, it considers the dynamic impact of each retrieved sample based on the LLM's feedback. Our ranking system system raises the ranks of more informative and stable samples and lower the ranks of misleading samples efficiently. Evaluations of various QA datasets from different domains show that our proposed method outperformed all the baselines, and especially improve the LLM's prediction on the long-tail questions.
 \section{Limitations}
%we ignore the order of the selected samples
There are several limitations of our work.

First, our method do not consider the effect of different orders within the retrieved set and rank the retrieved samples according to their ranking scores. Future works can be extended based on our work by considering different inner order within the retrieved set and their effect on the prediction results. 
%cannot work for those do not have related samples in the candidate pool

Second, although our experimental results show that our method greatly improves the prediction accuracy on long-tail samples, our method cannot handle query cases with no related knowledge either in the pre-training set or candidate pool.   %other tasks summarization, translation, recommendation

Third, our method focused on QA tasks using LLM. For future work, our method can be extended to other tasks such as summarization, translation, and recommendation as follows. Since our method is to train a reranker based on the reward signal from LLM, to adapt to other tasks, we can modify the evaluation score that is used to determine the reward. If the accuracy of the LLM’s predicted answer is unavailable, alternative metrics such as BLEU and ROUGE can be used to assess the consistency between the prediction and the ground truth. A threshold can then be set for these scores, where values exceeding the threshold yield a positive reward, while lower values result in a negative reward.

%\section*{Acknowledgments}

\bibliography{custom}
\clearpage
\appendix
\begin{table*}[ht]
    \centering
    \begin{tabular}{cccccc}
    \toprule
        Dataset &Type& Domain&Training&Test&Prompt format \\
         \hline
        Pubmedqa&Multi-choice&Healthcare &1000 &500&SQO-A\\
        ethos-national&Multi-choice&Speech detection&476&298&QO-A\\
        eval-climate&Multi-choice&climate change&288&180&QO-A\\
        T-REx&Open-ended&Wikipedia&20128&5032&Q-A\\
        NatQA&Open-ended&Wikipedia&11476&2869&Q-A\\
         \bottomrule
    \end{tabular}
    \caption{The statistics of the datasets used in this paper.}
    \label{tab:dataset}
\end{table*}
\begin{table*}[ht]
    \centering
    \resizebox{0.8\textwidth}{!}{
    \begin{tabular}{cp{6cm}p{6cm}}
    \toprule
    Notation&Retrieval sample format& Query sample format\\
    \hline
        Q-A & Question: <question> Answer: The answer is <answer> &Question: <question> Answer:\\
        \hline
        QO-A & Question: <question>  Options: (A) <option A> (B) <option B> (C) <options C>... Answer: The answer is <answer>&Question: <question>  Options: (A) <option A> (B) <option B> (C) <options C>... Answer:\\
        \hline
        SQO-A&Statement: <context> Question: <question>  Options: (A) <option A> (B) <option B> (C) <options C>... Answer: The answer is <answer>&Statement: <context> Question: <question>  Options: (A) <option A> (B) <option B> (C) <options C>... Answer: \\
        \bottomrule
    \end{tabular}
    }
    \caption{Prompt Format notations.}
    \label{tab:prompt_format}
\end{table*}

\section{Appendix}
\subsection{Experiment Details}\label{sec:exp_detail}
\textbf{Dataset Details.}
In this paper, we evaluate across five QA datasets from different domains including multi-choice QA and open-ended QA. The detailed statistics of these datasets and the prompt format we used are shown in \cref{tab:dataset} and \cref{tab:prompt_format}. We conduct the train-test split for the last four datasets following~\citet{li2024does}. We randomly sample $1000$ samples from the training dataset if the training set size exceeds $1000$ to simulate the scenario where only a limited number of samples can be collected.
\subsection{Extended Experimental Results}
\textbf{More case study on the uncertainty of ICL.} According to \cref{tab:extended_uncertian_case} on the healthcare dataset Pubmedqa, LLM can achieve correct prediction with the first two retrieved samples but gives a wrong prediction when the third sample is added to the prompt, which indicates that the third sample is misleading.

\textbf{More case study on hard samples.}
\cref{tab:hard_case_trex} shows another case for retrieved samples of hard samples on T-REx. According to the results, the query question asks about the instance of a subject, while the prompt retriever retrieved samples about the questions related to the locations, which mislead the final prediction. For our retriever, all the retrieved samples are related to the questions related to the instance of the subject, and provide informative augmentations for the inference.

\begin{table*}[]
    \centering
    \resizebox{1\textwidth}{!}{
    \begin{tabular}{p{3.5cm}p{14cm}|c}
    \toprule
    \multicolumn{3}{c}{\textbf{Query:} Statement: Lymphedema may be identified by... Question: Can a practicing surgeon detect early lymphedema reliably?}\\
    \hline
         \multicolumn{2}{c}{Retrieved samples}&Prediction  \\
         \hline
        Retrieved sample 1&Statement: Minority patients with cancer experience... Question: Can patient coaching reduce racial/ethnic disparities in cancer pain control? Answer: Yes.& Maybe (\Checkmark) \\
        \hline
        Retrieved sample 1 + sample 2&Statement: Minority patients with cancer experience... Question: Can patient coaching reduce racial/ethnic disparities in cancer pain control? Answer: Yes. Statement: The potential effects of binge drinking during pregnancy... Question: Does binge drinking during early pregnancy increase the risk of psychomotor deficits? Answer: No.& Maybe (\Checkmark)\\
        \hline
        Retrieved sample 1 + sample 2 + sample 3&Statement: Minority patients with cancer experience... Question: Can patient coaching reduce racial/ethnic disparities in cancer pain control? Answer: Yes. Statement: The potential effects of binge drinking during pregnancy... Question: Does binge drinking during early pregnancy increase the risk of psychomotor deficits? Answer: No. Statement: Despite the advantages from using aromatase inhibitors... Question: Do adjuvant aromatase inhibitors increase the cardiovascular risk in postmenopausal women with early breast cancer? Answer: Yes.&No (\XSolidBrush)\\
         \bottomrule
    \end{tabular}
    }
    \caption{Extended case study for the uncertainty of ICL on Pubmedqa.}
    \label{tab:extended_uncertian_case}
\end{table*}

\begin{table*}[t]
    \centering
    \resizebox{0.8\textwidth}{!}{
    \begin{tabular}{c|p{6cm}|p{6cm}}
    \toprule
    Query question&\multicolumn{2}{c}{Outlaw [SEP] instance of.}\\
    \hline
       Retriever&PromptPG retriever  & Our retriever \\
       \hline
         \multirow{14}{*}{Retrieved samples} &Question: Hingani Dam [SEP] country. Answer: The answer is India.&Question: Schleich [SEP] instance of. Answer: The answer is municipality of Germany.\\
        \cline{2-3}
         &Question: Maryland State Archives [SEP] applies to jurisdiction. Answer: The answer is Maryland.&Question: Chevry-sous-le-Bignon [SEP] instance of. Answer: The answer is commune of France.\\
         \cline{2-3}
         &Question: Silvia Panguana [SEP] country of citizenship. Answer: The answer is Mozambique.&Question: The Listel Hotel [SEP] instance of.
Answer: The answer is hotel.\\
\cline{2-3}
         &Question: New Paluvayi [SEP] located in the administrative territorial entity. Answer: The answer is Andhra Pradesh.&Question: Westona [SEP] instance of. Answer: The answer is railway station.\\
        \cline{2-3}
         &Question: The '59 Sound [SEP] country of origin. Answer: The answer is United States of America.&Question: Secu [SEP] instance of. Answer: The answer is commune of Romania.\\
         \hline
         Prediction& film. (\XSolidBrush)&wooden roller coaster. (\Checkmark)\\
         \bottomrule
    \end{tabular}
    }
    \caption{Extended case study for retrieved samples of hard samples on T-REx.}
    \label{tab:hard_case_trex}
\end{table*}

\end{document}